\def\year{2018}\relax
\documentclass[letterpaper]{article} 
\usepackage{aaai18}  
\usepackage{times}  
\usepackage{helvet}  
\usepackage{courier}  
\usepackage{url}  
\usepackage{graphicx}  

\usepackage{amsmath}
\usepackage{amssymb}
\usepackage{gensymb}
\usepackage{algorithm}
\usepackage{enumerate}
\usepackage{subcaption}
\usepackage{caption}
\usepackage{paralist}
\usepackage{gensymb}
\usepackage{color}
\usepackage{soul}
\usepackage{bm}
\usepackage[printwatermark]{xwatermark}
\usepackage{xcolor}
\usepackage{booktabs}

\newcommand{\etal}{\emph{et al.}\ }

\begin{document}
%
\title{RMPD -- A Recursive Mid-Point Displacement Algorithm for Path Planning}
\author{Fangda Li \and Ankit V. Manerikar \and Avinash C. Kak \\
School of Electrical and Computer Engineering, Purdue University \\
465 Northwestern Avenue \\
West Lafayette, Indiana 47907 \\
\{li1208, amanerik, kak\}@purdue.edu}

\maketitle
\begin{abstract}
Motivated by what is required for real-time path planning,
the paper starts out by presenting RMPD, a new recursive
``local'' planner founded on the key notion that, unless
made necessary by an obstacle, there must be no deviation
from the shortest path between any two points, which would
normally be a straight line path in the configuration space.
Subsequently, we increase the power of RMPD by introducing
the notion of cost-awareness into the algorithm to improve
the path quality -- this is done by associating obstacle and
smoothness costs with the currently selected path points and
factoring those costs in choosing the best points for the
next iteration.  In this manner, the overall strategy in the
cost-aware form of RMPD, cRMPD, combines the computational
efficiency made possible by the recursive RMPD planner with
the cost efficacy of a stochastic trajectory optimizer to
rapidly produce high-quality local collision-free paths.
Based on the test cases we have run, our experiments show
that cRMPD can reduce planning time by up to two orders of
magnitude as compared to RRT-Connect, while still
maintaining a path length optimality equivalent to that of
RRT*.
\end{abstract}

\section{Introduction}

Path planning has been an important area of research in
robotics over the last several decades.  Path planning
algorithms have important uses in robotic assembly
\cite{nof1999handbook}, autonomous driving
\cite{kuwata2009real}, kinematic and dynamic control for
robots, etc.

In traditional approaches to path planning, one first
overlays a grid of points on the configuration space and
then develops an obstacle-free path incrementally from the
start configuration to the goal configuration, going from
one grid point to the next in the process.  Subsequently, a
search is carried out over the paths thus discovered to find
the optimal path that connects the goal configuration with
the start configuration.  These algorithms are known to work
well in low-dimensional configuration spaces.  However, as
the dimensionality of the space increases, they extract a
large performance penalty.

\begin{figure}
\centering
\includegraphics[width=0.3\textwidth]{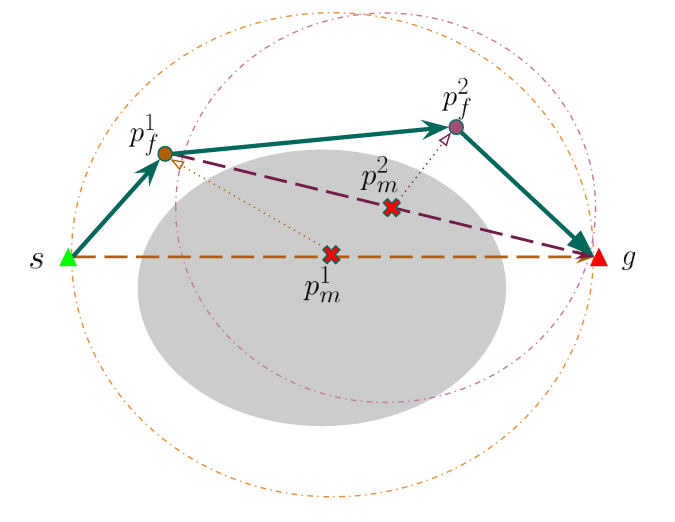}
\caption{RMPD uses recursion to efficiently find a detour
  around an obstacle.  Shown as $p_m^1$ and $p_m^2$ are the
  mid-points of two candidate path segments, each of which
  contains at least one in-collision point. The detouring
  collision-free replacements for the mid-points are shown
  as $p_f^1$ and $p_f^2$. The dashed circles represents the
  search hypersphere for middle point replacement.}
\label{fig:srmpd}
\end{figure}

More recently, sampling based algorithms have gained
considerable prominence.  The basic idea of such algorithms
is to sample the configuration space at randomly selected
points until a connection of the local pathways thus
constructed can lead one from the start configuration to the
goal configuration. These are best exemplified by the
Probabilistic Roadmap Method
(PRM) \cite{kavraki1996probabilistic}, the Randomized
Potential Fields \cite{barraquand1993nonholonomic}, and the
Rapidly-exploring Random Tree
(RRT) \cite{lavalle1998rapidly}.  The
important factors that account for the popularity of such algorithms include 
: (1)
They can be used with greater ease in high dimensional
configuration spaces compared to the traditional algorithms;
(2) The algorithms based on PRM and RRT can be shown to be
probabilistically complete; (3) The ease in merging
partially developed solutions in order to respond
simultaneously to multiple path planning queries; etc.

In our laboratory we have been exploring the use of RRT and
RRT-Connect \cite{kuffner2000rrt} algorithms for real-time
motion planning as needed for automatic pruning of dormant
trees --- a crucial step in growing healthy apple orchards.
Our main concern has been the limited extent to which a
straightforward application of RRT-Connect, as originally
formulated by its authors, can generate smooth and concise
paths in obstacle-dense regions.

To address this shortcoming of the randomized algorithms, we
propose a new approach here.  The first part of our approach
consists of a novel ``local'' planner that is based on the
key notion that, unless made necessary by an obstacle, there
must be no deviation from the shortest path between any two
points, which would normally be a straight line path in the
configuration space.  We refer to this basic path planner as
the RMPD algorithm or the single-path ``Recursive Mid-Point
Displacement'' algorithm.  Its basic idea is recursive:
Check each sampling point on the path connecting two
end-points for being collision free. If that condition is
not satisfied at any sampling point, move the mid-point to a
collision-free location to create a detour, divide the
resulting path into sub-paths and attempt to find a
collision-free detour for each such sub-path.  If the detour
is again in collision, further divide the sub-path, and so
on.

Subsequently, we increase the power of RMPD by including
cost-awareness in the algorithm. This is done by modifying
the basic search strategy of RMPD in a manner similar to
that of STOMP (Stochastic Trajectory Optimization Motion
Planning) \cite{kalakrishnan2011stomp} in order to steer the
iterative sampling distributions towards cost-optimal
regions in the sampling space. As a consequence, the
cost-aware RMPD, which we denote cRMPD, not only optimizes
for computational efficiency during path-planning but does
so with a degree of cost-awareness in producing
collision-free paths between the source and the goal points.

Following \cite{kalakrishnan2011stomp}
and \cite{ratliff2009chomp}, the cost function incorporated
in the cRMPD planner consists of an obstacle cost calculated
from a local potential field as well as a smoothness cost to
avoid rapid accelerations.  This cost is calculated for each
sampled point in the current iteration and is used to update
the sampling distribution in the subsequent search.  It
would seem that, while one can expect this cost-function
weighted sampling strategy to improve the cost efficiency of
the planner, it would do so by slowing down the planner.  It
is interesting to note, however, that the overall time
efficiency of the cRMPD planner does not suffer because the
search for obstacle-free paths is more likely to be
biased towards collision-free regions of the configuration
space.  That search bias results in the final paths to be
discovered in a shorter time. We will demonstrate this effect
through our experimental results.

In the experimental section, we demonstrate our algorithm
using both 2D bitmap and 3D environments with narrow
passages that must be traversed by a path from the start
state to the goal state. In addition, we evaluate the
performance of cRMPD using a simulation of apple tree
pruning. Based on the test cases we have run, we observe
that cRMPD significantly reduces the average path length
and also the control effort required.
Furthermore, cRMPD reduces the planning time by two orders of
magnitude in some of the test cases vis-a-vis the other
algorithms chosen for comparison.

In the rest of the paper, Section \ref{sec:rw} presents the
related work that is most relevant to cRMPD.  
In particular,
this section provides an overview of sampling based algorithms along with the
related past research that has focused on addressing the
shortcomings of these algorithms in exploring  narrow passages as
well as their cost-effectiveness.
Subsequently,
Section \ref{sec:rmpd} presents our cRMPD algorithm.
Section \ref{sec:exp} presents the experimental results that
demonstrate the efficiency of the cRMPD algorithm vis-a-vis
its main competition.  Finally, we conclude in
Section \ref{sec:con}.

\section{Related Work}
\label{sec:rw}

We start with an overview of the sampling based algorithms for path planning.
Such algorithms are generally either
graph-based or tree-based.  Graph based motion planning
algorithms, with Probabilistic Roadmap Method
\cite{kavraki1996probabilistic} being the most
representative, are widely used in multi-query motion
planning scenarios. A 
``roadmap'' only needs to be constructed once by such planners for static workspaces; the same roadmap
can subsequently be used to handle multiple
queries. On the other hand, tree based algorithms, such as
the Rapidly-exploring Random Tree (RRT)
\cite{lavalle1998rapidly}, build space filling trees with
biased growth toward large unexplored regions.

From the standpoint of efficiency, a problem with randomized
sampling based algorithms, such as RRT, is that they tend to
carry out excessive unnecessary sampling as only a tiny
fraction of what is explored in the configuration space is
eventually used for constructing the solution path.  Many
approaches have been proposed to address this problem.  For
example, for paths in the vicinity of narrow passages, one
can increase the odds of a path traversing those passages by
directly biasing the sampling to favor obstacle-dense
regions
\cite{lee2012sr,hsu2003bridge,zhang2008efficient,urmson2003approaches}
or by actively reducing the sampling in wide-open free
spaces
\cite{lee2012sr,urmson2003approaches,kuffner2000rrt,clifton2008evaluating}. The first strategy is referred to by
``Narrow-passage Favored Sampling'' and the second by
``Open-Space Selective Sampling'', respectively.  In what
follows, we will provide brief reviews of these two
strategies.

\textbf{Narrow-passage Favored Sampling} aims 
explicitly to sample more densely in potentially narrow
regions of the configuration space.  In order to identify
such regions, Hsu \etal has proposed a {\em bridge test} in
which a bridge is defined as a line segment consisting of
two in-collision endpoints and one free middle
point \cite{hsu2003bridge}.  Other approaches such as
by \cite{zhang2008efficient} propose a retraction-based
modification to RRT, which generates dense samples near
obstacles at the cost of a higher computational overhead.
Retracting a sample in this context refers to the process of
resampling around the sample with a bias to improve upon the
current choice of the sample with respect to a certain
metric.  The contribution reported in \cite{lee2012sr}
attempts to integrate both these methods by retracting the
samples that pass the bridge test.

On the other hand, the main idea of \textbf{Open-space
Selective Sampling} is to implicitly bias the sampling
toward narrower regions in the configuration space by
biasing it against wide-open regions.  An approximate
approach for doing so is proposed in \cite{lee2012sr} in
which a free hypersphere is associated with every
non-contact RRT node and samples falling within any of the
free hyperspheres are discarded.  In the bidirectional RRT
algorithm, namely RRT-Connect \cite{kuffner2000rrt}, a
greedy path extension function decreases the need for
excessive sampling in the open regions of the configuration
space.  Moreover, with the same path extension logic, an
arbitrary number of RRTs can be grown simultaneously to
increase the pace of exploration in narrow regions, as
described in \cite{clifton2008evaluating}.

Our algorithm incorporates both strategies mentioned above
in an intuitive manner.  By aggressively connecting samples
with straight lines, whenever that is possible, RMPD
significantly diminishes the over exploration tendency of
RRT. At the same time, RMPD explores more densely in the
vicinity of narrow regions through the detouring logic
triggered by in-collision samples.

In general, despite being probabilistic complete, the
traditional formulations of sampling based algorithms remain
agnostic regarding the costs associated with the paths.
Fortunately, it has been known for some time that it is
possible to include cost considerations in sampling based
approaches to motion planning
\cite{karaman2011sampling}.
While the contribution in \cite{karaman2011sampling} can serve as a
guide to combining cost considerations with sampling based planners,
the paths produced still need post-processing in order to be viable.
In that context, it is interesting to note that the trajectory optimization
based approach as incorporated in CHOMP \cite{ratliff2009chomp} requires
no such post-processing.  
CHOMP uses Hamiltonian cost optimization to generate smooth
motion paths.
However, the gradient descent method 
utilized in CHOMP often result in entrapment in local minima. 

To get around the problems caused by getting trapped in
local minima, Kalakrishnan \etal developed the STOMP planner
that uses stochastic sampling to search for
cost optimal paths \cite{kalakrishnan2011stomp}.
We use this method to transform RMPD into the cost-aware algorithm,
cRMPD.
cRMPD combines the time-efficiency of
sampling-based path search with the
cost-effectiveness of an optimal-control planner to rapidly
generate local cost-aware paths for solving motion planning
problems.
In Section \ref{sec:exp}, we compare the performance of cRMPD
with the other algorithms mentioned in this section.  

\section{RMPD -- Recursive Mid-Point Displacement Algorithm}
\label{sec:rmpd}

In this section, we propose a novel path planning algorithm
we call RMPD, for Recursive Mid-Point Displacement, that is
based on a simple path construction idea: Don't deviate from
a straight-line unless absolutely necessary on account of
the obstacles.

We first present the base version of the algorithm, 
which focuses on overcoming local obstacles
without much computational overhead.  In order to find a
path, RMPD connects any two nodes with a straight line,
which is checked for in-collision property using a local
planner.  If this property is found to be true at any
sampling point on the straight-line path, tangential detours
around the obstacle are made recursively by replacing the
line's middle point.  

Subsequently we substitute the collision-free middle
point search in RMPD with a cost-aware exploration.  The
derived algorithm, cRMPD, thereby inherits the time
efficiency of RMPD and further improves the path quality by
incorporating a numerical cost function.

\subsection{Single-path RMPD}

In contrast to the retraction methods described in
\cite{zhang2008efficient,lee2012sr}, our basic algorithm,
RMPD, navigates its way around obstacles with inexpensive
resampling, while leveraging the idea of divide and conquer.
As the recursion goes deeper, RMPD focuses on finding a
warped collision-free replacement for a shorter path segment
within its local area.

The steps of RMPD are presented in Algorithm \ref{al:rmpd}.
Using the notation shown in Figure \ref{fig:srmpd}, given
a start point $p_s$ and a goal point $p_g$ in a
configuration space, the algorithm first validates $p_s$ and
$p_g$ and returns with failure if either $p_s$ or $p_g$ is in
collision.  
If both states are valid, RMPD invokes itself recursively to populate an 
initially empty path $\boldsymbol{\theta}$ rooted at $p_s$.
If the condition returned by the terminating recursion is $true$, we have found a
valid path $\boldsymbol{\theta}$.

\begin{algorithm}
  \floatname{algorithm}{Algorithm}
  \caption{$RMPD(\boldsymbol{\theta},p_{s},p_{g})$}
  \begin{enumerate}[1]
	\item \textbf{if} $p_s$ or $p_g$ is in collision
    \item \quad \textbf{return false}
    \item \textbf{if} $\overline{(p_{s},p_{g})}$ is in collision
    \item \quad $p_{m} \Leftarrow (p_{s}+p_{g})/2$
    \item \quad \quad \textbf{if} $p_{m}$ is in collision
    \item \quad \quad \quad $\sigma \propto \left \|p_s - p_g \right \|$
    \item \quad \quad \quad $p_{f} \Leftarrow GaussianFreeStateSampler(p_{m}, \sigma)$
    \item \quad \quad \textbf{else}
    \item \quad \quad \quad $p_{f} \Leftarrow p_{m}$
    \item \quad \textbf{return} $RMPD(\boldsymbol{\theta},p_{s},p_{f})$ and $RMPD(\boldsymbol{\theta},p_{f},p_{g})$
    \item \textbf{else}
    \item \quad $\boldsymbol{\theta}.append(p_{g})$
    \item \quad \textbf{return} \textbf{true}
  \end{enumerate}
  \label{al:rmpd}
\end{algorithm}

As a key step, RMPD recursively breaks a path segment into
two potentially collision-free sub-paths and appends the
middle point to the final path, with the main consideration
behind employing the middle point for displacement being
simplicity.  In line 3 of Algorithm \ref{al:rmpd}, RMPD
first attempts to connect $p_{s}$ and $p_{g}$ directly with
a straight line $\overline{(p_{s},p_{g})}$ using a
bidirectional local planner, which validates the segment by
marching from the middle point to the two ends in an
alternating fashion.  If no obstacle is encountered, the
function simply adds $p_{g}$ to $\boldsymbol{\theta}$ as a
waypoint in the final path (line 12).  On the other hand, if
a path segment $\overline{(p_{s},p_{g})}$ encounters an
obstacle at any sampling point, RMPD tries to make a
tangential detour around the obstacle by breaking
$\overline{(p_{s},p_{g})}$ into two halves.

First, $p_{m}$, the middle point of $p_{s}$ and $p_{g}$, is
checked for collision.  In case of collision, $p_{m}$ is
replaced by a nearby collision-free point $p_{f}$, returned
by $GaussianFreeStateSampler(p_{m}, \sigma)$ (line 7).  The
standard deviation of the Gaussian sampler is set
proportional to the distance between $p_{s}$ and $p_{g}$,
which ensures locality of the search for a free middle
point.  
If the maximum number of iterations is reached
before the discovery of a free point, the sampler simply
returns its last failed candidate, which causes the validity check in the next level of recursion to fail (line 1).
Finally, RMPD recursively invokes itself with both
sub-paths, $\overline{(p_{s},p_{f})}$ and
$\overline{(p_{f},p_{g})}$ (line 10).  Recursion terminates
automatically when the waypoint states in
$\boldsymbol{\theta}$ connect $p_{s}$ to $p_{g}$
successfully with consecutive collision-free path segments.

In the implementation of RMPD, we specify the max number of
waypoints in the path, $N_{max}$.  If the initial $p_s$ and
$p_g$ are still not connected after having $N_{max}$
waypoints in $\boldsymbol{\theta}$, we deem that RMPD
invocation to have failed to find a solution.

\subsection{Cost-Aware RMPD (cRMPD)}

While the time efficiency of a planner is desirable in real-time
planning scenarios, path quality is yet another important
discriminating factor.  In this section, we introduce a
cost-aware modification to the search strategy for the
middle point in RMPD as implemented by the call to
$GaussianFreeStateSampler$ in Line 7 of
Algorithm \ref{al:rmpd}.  As opposed to keeping only the
first feasible sample in the basic RMPD, cRMPD computes the
cost of a sampled point regardless of its validity and then
aggregates the cost information with those of other nearby
points for the purpose of choosing the best candidate for the
next middle point.

cRMPD is implemented by replacing the call to
$GaussianFreeStateSampler$ in Line 7 of
Algorithm \ref{al:rmpd} by a call to
$CostAwareFreeStateSampler$ whose pseudo code is shown in
Algorithm \ref{al:crmpd}.  Note that
$CostAwareFreeStateSampler$ includes a call to
$GaussianSampler$ in line 8 of Algorithm \ref{al:crmpd}.
For the difference between $GaussianFreeStateSampler$ used
in RMPD and $GaussianSampler$ used in cRMPD, the former
carries out repeated sampling until a collision free point
is found, whereas the latter returns a sample regardless of
its collision property.

In every iteration, the Gaussian sampler ($GaussianSampler$)
with mean $p_m$ and standard deviation $\sigma$ first
generates $K$ random samples (lines 7 and 8).  Similar to
what is done in RMPD, the standard deviation $\sigma$ of the
Gaussian sampler is proportional to the straight line
length.  Subsequently, the differences between the random
samples and the current middle point are added together
using exponentiated weights to produce the update $\delta p$
(line 10).  The sampling distribution in the next iteration
is centered at the updated middle point, which is
essentially the weighted average of the previous K samples.
The weights are computed with the softmax function that is
based on the cost of each sample $f(p_i)$ (line 9), where
$h$ is a constant.  These weights can also be interpreted as
probabilities used in calculating the expectation of the
true gradient.  Using the exponentiated weights as shown
amounts to implementing the EGD (Estimated Gradient Descent)
method presented in \cite{kalakrishnan2011stomp}.  The
search for the optimal middle point using this approach is
repeated until either the maximum number of attempts is
reached or the cost converges.

\begin{algorithm}
  \floatname{algorithm}{Algorithm}
  \caption{$CostAwareFreeStateSampler(p_{s},p_{m},p_{g})$}
  \begin{enumerate}[1]
	\item $c \Leftarrow +\infty$
    \item $\delta p \Leftarrow 0$
    \item $\sigma \propto |p_s - p_g|$
    \item \textbf{do}
	\item \quad $c_{prev} \Leftarrow c$
    \item \quad $p_m \Leftarrow p_m + \delta p$
    \item \quad \textbf{for} $i = 1, ..., K$
    \item \quad \quad $p_i \Leftarrow GaussianSampler(p_m, \sigma)$
    \item \quad \quad $w_i \Leftarrow \frac{e^{- h f(p_i)}}{\sum^K_j e^{- h f(p_j)}}$
    \item \quad $\delta p \Leftarrow \sum^K_i w_i (p_i - p_m)$
    \item \quad $c \Leftarrow f(p_m)$
    \item \textbf{while} $c_{prev} - c > \epsilon$
    \item \textbf{return} $p_m$
  \end{enumerate}
  \label{al:crmpd}
\end{algorithm}

In a manner similar to STOMP, through the loop in lines 4
through 12, cRMPD employs a Monte Carlo approach for the
cost function minimization.  Moreover, if one regards the
problem of finding an optimal middle point as a special case
of STOMP with only three waypoints, the random exploration
strategy on trajectories by adding multivariate Gaussian
noise is then analogous to Gaussian sampling for
discovering the best middle point.

While cRMPD can accommodate any arbitrary cost function, in
the interest of computational efficiency in a sampling-based
context, the cost function $f(p)$ should be only state
dependent and should be computable in constant time.
Following the cost functions often used in trajectory
optimization
algorithms \cite{ratliff2009chomp,kalakrishnan2011stomp},
the cost function in cRMPD consists of a clearance term and
a smoothness term:
\begin{equation}
f(p) = f_{clr}(p) + \lambda f_{smt}(p).
\end{equation}
The clearance term, necessary for finding tangential detours
around the obstacles, is taken from a signed distance field
in the configuration space.  The magnitude of $f_{clr}(p)$
represents the distance from $p$ to the closest point of
opposite in-collision property\footnote{What that means is
that if $p$ is in-collision, meaning it is inside an obstacle,
then we seek the closest point that is collision free. On
the other hand, if $p$ is collision-free, we seek that point
that is on the nearest obstacle.}:
\begin{equation}
f_{clr} (p) = sgn(p) \cdot \min \{|p - p_i| \mid p_i \in P_{-sgn(p)}\}.
\end{equation}
The polarity function $sgn(p)$ returns either $1$ or $-1$
such that $f_{clr}$ is positive when $p$ is in collision.

The smoothness term is defined as:
\begin{equation}
f_{smt} (p) = |p_s - p| + |p - p_g| - |p_s - p_g|.
\end{equation}
Basically, $f_{smt}$ punishes any middle point that deviates
from the original straight line path, and thereby
incentivizes shorter and smoother paths.

The cost-aware sampling strategy of the cRMPD planner allows
it to be likened to the STOMP
planner \cite{kalakrishnan2011stomp}, albeit with one main
distinction. The distinguishing factor between the cRMPD and
STOMP planner is the number of waypoints accounted for
during the iterative sampling -- a STOMP planner samples a
new set of $n$-point trajectories for each iteration, where
$n$ is fixed throughout the path planning operation. On the
other hand, a cRMPD planner recursively samples path points
to replace the middle-point of a path segment to form a
collision-free path. 
Thus, whereas $CostAwareFreeStateSampler$ can be considered as a special case of STOMP in which the number of waypoints
is fixed to be three, in cRMPD the total number of waypoints
increases with each iteration as required by the
planner. 
This flexibility in the number of waypoints therefore
allows cRMPD to adjust discretization of the path to suit
the complexity of the workspace.

\section{Experiments}
\label{sec:exp}

In this section, we compare the performance of cRMPD with
other sampling-based algorithms, namely
RRT \cite{lavalle1998rapidly},
RRT-Connect \cite{kuffner2000rrt},
RRT* \cite{karaman2011sampling} and
PRM \cite{kavraki1996probabilistic}.  The algorithms are
benchmarked on three planning problems: point robot on a 2D
bitmap, the Piano Mover's problem, and the Twistcooler
problem.  The planning scenarios are chosen to be rich in
narrow passages such as the middle hole in Twistcooler and
the narrow corridor in Piano Mover.  The 
maximum path-planning time was set to be
sufficiently high to allow RRT* to converge
and a near-optimal cost path could be found.  We show that    
cRMPD achieves a more favorable trade-off between time
efficiency and path quality than RRT and its popular
variants.  Subsequently, the performance of cRMPD is
evaluated against the trajectory optimizer STOMP in a tree
pruning benchmark.  We demonstrate that cRMPD is capable of
generating smooth paths with quantitative metrics comparable
to those by the trajectory optimizer, while using much less
time.

\subsection{Experimental Setup}

We have implemented cRMPD in C++ within the
OMPL \cite{sucan2012open} framework.  OMPL also provides
optimized implementations for the competing sampling-based
algorithms --- this ensures fairness when comparing
experimental results.  For the tree pruning benchmark, we
used the publicly available STOMP
implementation \cite{stompmoveit} within the
MoveIt! \cite{moveit} framework.  All experiments were
repeated 30 times on a 2.30 GHz Intel i7 processor with 8 GB
of RAM and the average results are presented.

For all the paths produced by sampling-based planners,
including cRMPD, all quantitative measurements are extracted
after post-processing.  This post-processing includes
iterative path shortcutting followed by B-spline fitting per
the OMPL implementation.  Although the paths produced by the
different planners in our comparative evaluation are more or
less similar with and without post-processing, we only
present the results on smoothed paths since post-processing
is a crucial step for any geometric planner to generate
physically executable paths \cite{elbanhawi2014sampling}.
As long as this post-processing step is identical for all
the planners in a comparative evaluation, no one planner
gets any advantage.  Additionally, we do the following to
compute the smoothness of a path $\boldsymbol{\theta}$:
\begin{enumerate}
\item
Upsample the path uniformly to $M$ waypoints, so that every
two consecutive waypoints are equally spaced in the
configuration space $\mathbb{R}^n$.
\item
Apply an $M \times M$ second-order finite differencing
matrix $A$ to the interpolated path
$\boldsymbol{\theta}' \in \mathbb{R}^{M \times n}$
(Equation \ref{eq:acc}).
\item
Sum the norms of the $M$ vectors in $\boldsymbol{\theta}'$
to obtain the final smoothness value $q_{smt}$.
\end{enumerate}
\begin{equation}
\boldsymbol{A} = \begin{bmatrix}
1 & 0 & 0 &  & 0 & 0 & 0 \\ 
-2 & 1 & 0 & \cdots & 0 & 0 & 0 \\ 
1 & -2 & 1 &  & 0 & 0 & 0 \\ 
 & \vdots &  & \ddots &  & \vdots & \\ 
0 & 0 & 0 &  & 1 & -2 & 1 \\ 
0 & 0 & 0 & \cdots & 0 & 1 & -2 \\ 
0 & 0 & 0 &  & 0 & 0 & 1
\end{bmatrix}
\label{eq:fdm}
\end{equation}
\begin{equation}
\ddot{\boldsymbol{\theta}'} = \boldsymbol{A}\boldsymbol{\theta}
\label{eq:acc}
\end{equation}
Note that $q_{smt}$ measures the total amount of
acceleration or joint effort needed by the robot                
to traverse the path.

Lastly, as for the choice of the parameters $\lambda = 0.5$,
$h = 5.0$ and $N_{max} = 100$, these were determined
empirically.  Choosing $K$ carefully is critical to the
success of the algorithm.  
As the value of $K$ increases,
one tends to get shorter and smoother paths but at the expense of longer planning time.
We noticed in our
experiments that the benefits of a larger value for $K$
saturate at around $K=10$. So we have set $K$ to this value
in all experiments with cRMPD.
Additionally, setting $\sigma$ in the $CostAwareFreeStateSampler$ to be 1/6 of the distance between the start and the goal (line 3 in Algorithm \ref{al:crmpd}) yielded good results and was used for all experiments.

\begin{figure*}[!ht]
  \centering
  \begin{subfigure}[b]{0.25\textwidth}
  	\includegraphics[width=\textwidth, height=3.5cm]{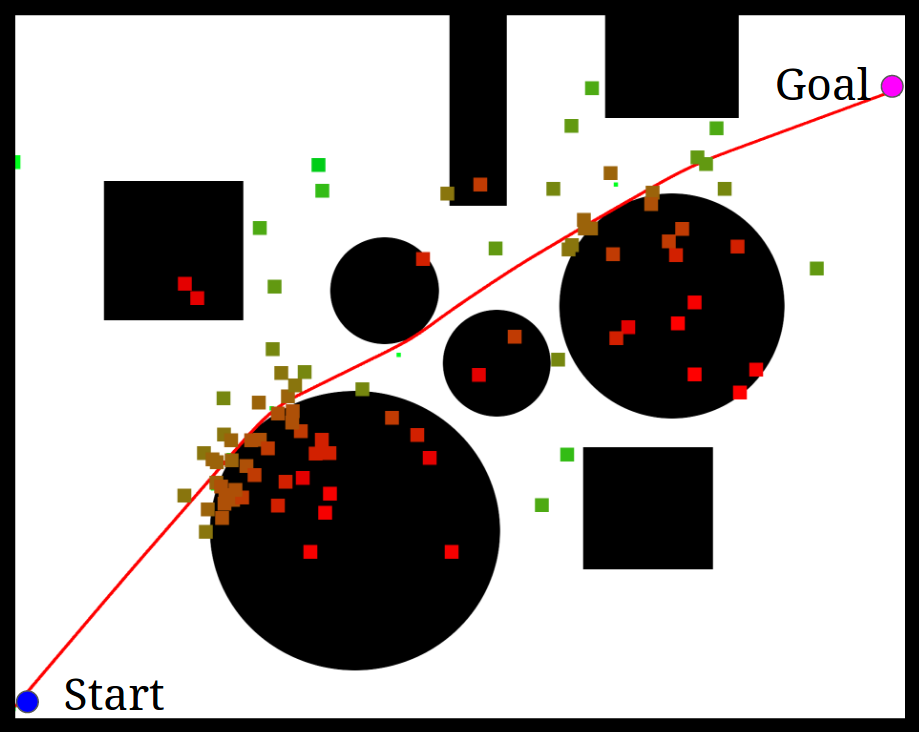}
  	\caption{}
  	\label{fig:bitmap}
  \end{subfigure} 
  \begin{subfigure}[b]{0.25\textwidth}
  	\includegraphics[width=\textwidth, height=3.5cm]{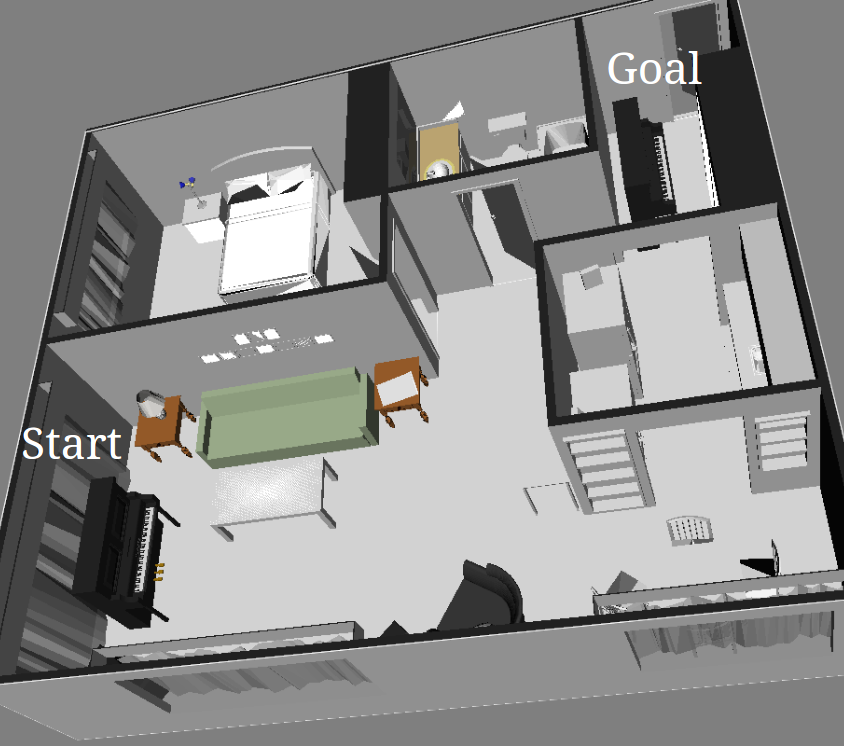}
  	\caption{}	
  	\label{fig:twistcool_env}
  \end{subfigure} 
   \begin{subfigure}[b]{0.25\textwidth}
  	\includegraphics[width=\textwidth, height=3.5cm]{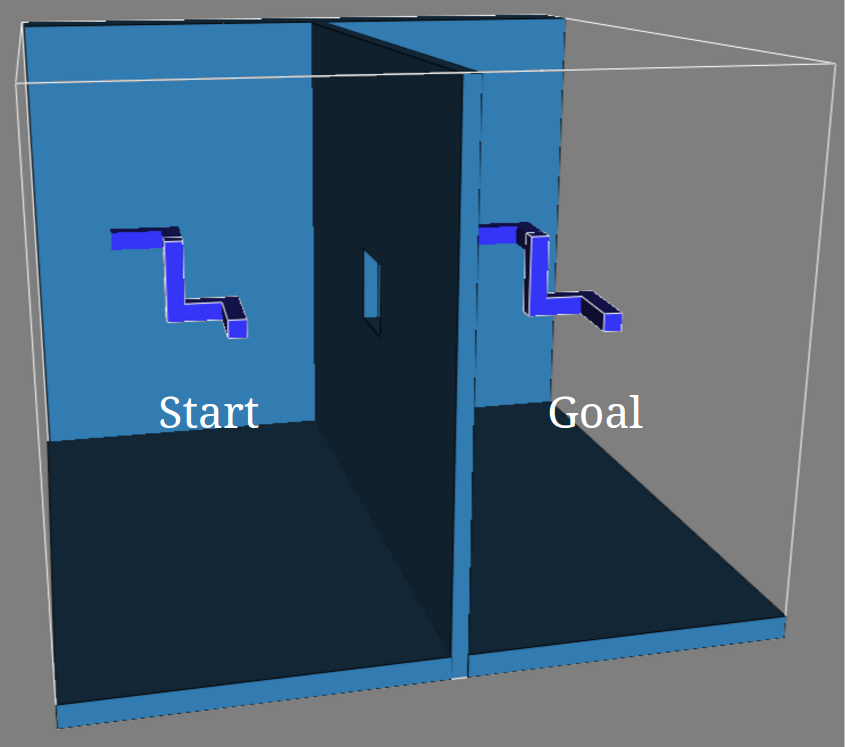}
  	\caption{}
  	\label{fig:piano_env}
  \end{subfigure} 
  
  \begin{subfigure}[b]{0.25\textwidth}
  	\includegraphics[width=\textwidth, height=5.3cm]{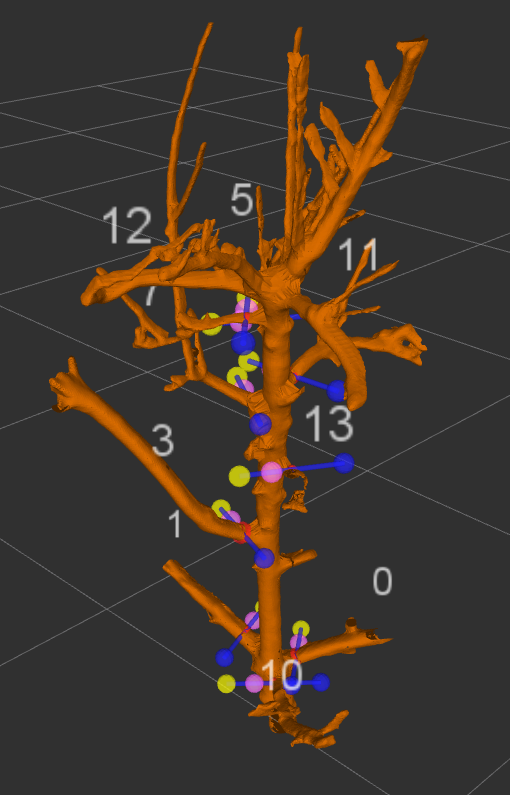}
  	\caption{}	
  	\label{fig:tree}
  \end{subfigure} 
  \begin{subfigure}[b]{0.25\textwidth}
  	\includegraphics[width=\textwidth, height=5.3cm]{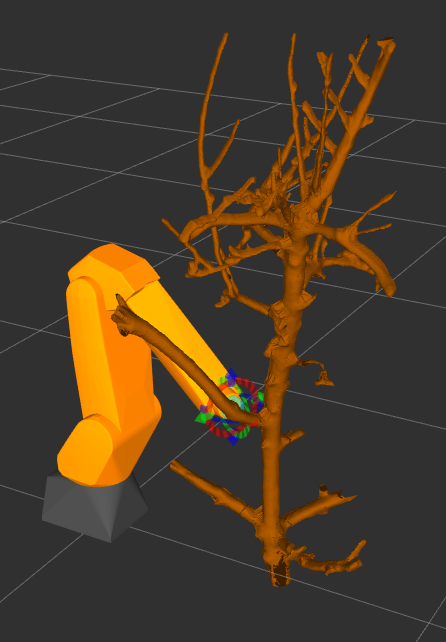}
  	\caption{}	
  	\label{fig:query}
  \end{subfigure} 
  \begin{subfigure}[b]{0.25\textwidth}
  	\includegraphics[width=\textwidth, height=5.3cm]{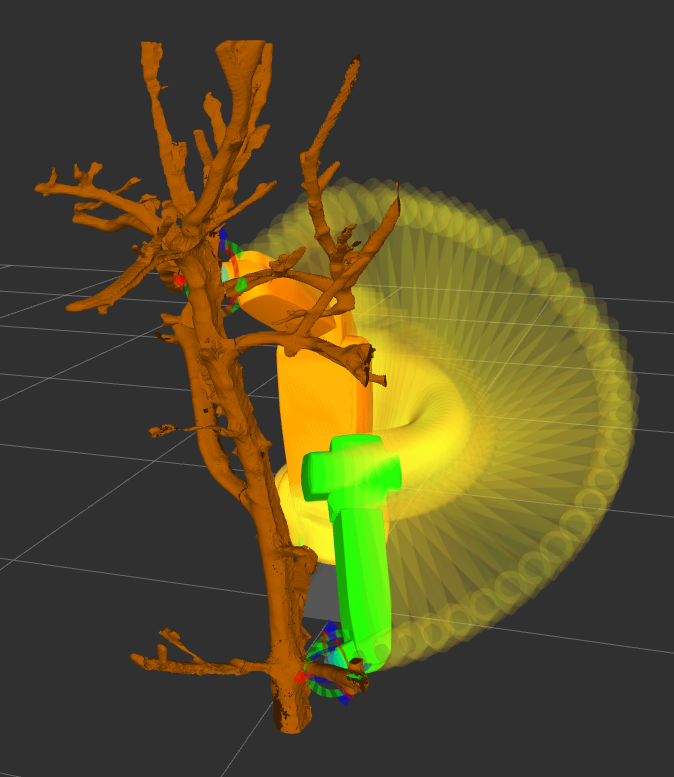}
  	\caption{}	
  	\label{fig:trail}
  \end{subfigure} 

  \caption{Spatial layouts for the four benchmarks and selected results. (a) A sample
  solution by cRMPD on the 2D bitmap. The approximated
  clearance cost of a sample is indicated by the color. (b)
  The Piano Mover's problem. (c) The Twistcooler
  Problem. (d) Mesh of the scanned apple tree. To prune a
  branch, the end effector must reach the pink marker while
  staying aligned to the blue arrow. (e) An example pruning
  pose used in the benchmark. (f) An example path found by
  cRMPD.}  \label{fig:benchmarks}
\end{figure*}

\subsection{Point Robot on 2D Bitmap}

The point robot, with 2 translational DoF,
in the bitmap case studied by us must navigate
from the bottom left corner of the map to the top right
corner, as shown in Figure \ref{fig:bitmap}.  The
environment contains a cluster of obstacles that are placed
to form a narrow direct passage to the goal along the
diagonal, while creating an open region near the bottom
right corner.  In this benchmark, every run is given a 0.5s
quantum, which is sufficient for the cost of RRT* to
converge.  Through the quantitative results in
Table \ref{tb:2D}, we observe that although RRT and
RRT-Connect score higher with regard to planning time, they
are likely to choose suboptimal path segments in the open
regions of the configuration space.  This observation is
best supported by the comparison on path lengths.  On the
other hand, navigating through the narrow diagonal passage
poses little challenge for the cRMPD planner.  With a
straight line initialization, cRMPD achieves a competitive
average path length comparable to what is returned by RRT*
although RRT* takes more than 10 times longer to find the
path.  Also note that our state-dependent cost function
allows cRMPD to save on the number of collision checks.

\begin{table}[tbp]
\centering
\caption{2D point robot benchmark results. Every column is normalized against the best performer to extract the relative performance (\# CC = number of collision checks).}
\label{tb:2D}
\begin{tabular}{@{}llll@{}}
\toprule
      & Time & Length & \# CC \\ \midrule RRT & 1.07 & 1.15
& 1.36 \\ RRT-C & \textbf{1.00} & 1.15 & 1.23 \\ RRT* &
427.34 & \textbf{1.00} & 43.28 \\ cRMPD & 31.84 & 1.03
& \textbf{1.00} \\ \bottomrule
\end{tabular}
\end{table}

\subsection{Path Planning in 3D Workspaces} 
To examine the characteristics of cRMPD in 3D workspaces, we
study two rigid-body motion planning problems.  The set of
transformations for all rigid-body motions in 3D workspaces
forms the special Euclidean manifold, $SE(3)$. This manifold
spans a 6-dimensional space consisting of two subspaces: a
3D Euclidean space for translations and another 3D space in
the special Orthogonal group $SO(3)$ that describes
rotations.

It is noteworthy that the distance between any two points on
the $SE(3)$ manifold is defined as a simple summation of
distances in the two subspaces: $\ell_2$-norm in the
Euclidean subspace and arc-length on the $SO(3)$
manifold \cite{kuffner2004effective}.  To avoid the
complications created by averaging the states in $SE(3)$, as
required by the EGD (estimated gradient descent) calculation in line 10
of Algorithm \ref{al:crmpd}, cRMPD adopts a greedy strategy
for choosing the middle point -- the updated middle point in
each iteration is simply the one with the lowest cost among
the $K$ samples.  As the reader will see shortly, this
works well in practice.

Of the two 3D path planning problems we first show results for the classic Piano Mover's problem,
as seen in Figure \ref{fig:piano_env}, where a piano must be
maneuvered through the living room and the narrow corridor
at the top right.  The results, presented in              
Table \ref{tb:piano}, show that cRMPD surpasses its
competitors by a significant margin with respect to nearly
all the metrics.  While being more than 10 times faster than
RRT-Connect, cRMPD still returns the shortest and smoothest
path, even compared to the optimal planner RRT*.

\begin{table}[tbp]
\centering
\caption{Relative performance for the Piano Mover's problem (SR = Success Rate and $q_{smt}$ = Smoothness Cost).}
\label{tb:piano}
\begin{tabular}{@{}llllll@{}}
\toprule
      & Time          & Length        & $q_{smt}$     & \# CC         & SR (\%)           \\ \midrule
RRT   & 32.74         & 1.25          & 8.75          & 71.46         & 80.0              \\
RRT-C & 15.17         & 1.22          & 7.78          & 38.55         & \textbf{100.0}    \\
RRT*  & 61.35         & 1.02          & 8.01          & 186.28        & 73.3              \\
PRM   & 37.62		  & 1.21		  & 9.20		  & 137.29 		  & 76.7              \\
cRMPD & \textbf{1.00} & \textbf{1.00} & \textbf{1.00} & \textbf{1.00} & 96.7              \\ \bottomrule
\end{tabular}
\end{table}

In the second 3D path planning problem, the Twistcooler problem, a 5-link
rigid-body robot must navigate through a small hole in a
barrier in the middle while there exist large open spaces on
the two sides of the barrier.  The Twistcooler problem is
deceivingly challenging: the small hole in the middle and
the stretched shape of the robot form a elongated narrow
passage in the configuration space.  For RRT and its
variants, a typical run lasts several hundreds of seconds
before a solution is found.  We did try to combine those
planners with obstacle-based samplers that generate samples
close to obstacles\footnote{An obstacle-based sampler first
takes two samples, one valid and the other invalid.
Subsequently, it interpolates from the invalid sample to the
valid sample, returning the first valid sample
encountered \shortcite{sucan2012open}.}, yet no major
improvement in planning time was observed.  The PRM planner
achieved the highest success rate and improved time
efficiency, however it did so with compromised path
smoothness and the most number of collision checks.

As shown by the benchmarking results tabulated in
Table \ref{tb:twist}, cRMPD possesses superior ability to
navigate through narrow passages.  Quantitatively speaking,
cRMPD is able to produce a path within a time span that is
one order of magnitude shorter than any other planner.  In
addition, cRMPD is able to do so without compromising the
path quality, as demonstrated the final solution length and
smoothness.

\begin{table}[tbp]
\centering
\caption{Relative performance for the Twistcooler problem.}
\label{tb:twist}
\begin{tabular}{@{}llllll@{}}
\toprule
      & Time          & Length        & $q_{smt}$     & \# CC         & SR (\%)           \\ \midrule
RRT   & 279.01        & 1.49          & 16.23         & 430.05        & 96.7              \\
RRT-C & 428.22        & 1.55          & 15.49         & 665.52        & 73.3              \\
RRT*  & 746.27        & \textbf{1.00} & 11.74         & 152.96        & 60.0              \\
PRM   & 52.81		  & 1.52		  & 21.62		  & 882.87  	  & \textbf{100.0}	  \\
cRMPD & \textbf{1.00} & \textbf{1.00} & \textbf{1.00} & \textbf{1.00} & 90.0              \\ \bottomrule
\end{tabular}
\end{table}

\subsection{Dormant Apple Tree Pruning}

\begin{table}[tbp]
\centering
\caption{Relative performance on the tree pruning benchmark.}
\label{tb:pruning}
\begin{tabular}{@{}lllll@{}}
\toprule
      & Time          & Length        & $q_{smt}$     & SR (\%)           \\ \midrule
RRT   & 17.28         & 1.47          & 2.16          & 91.5              \\
RRT-C & \textbf{1.00} & 1.12          & 15.49         & \textbf{100.0}    \\
RRT*  & 37.55         & 1.19          & 1.34          & 93.1              \\
STOMP & 24.75         & \textbf{1.00} & \textbf{1.00} & 41.5              \\
cRMPD & 1.81          & 1.07          & 1.68          & \textbf{100.0}    \\ \bottomrule
\end{tabular}
\end{table}

We now show results for a real-world problem: dormant apple
tree pruning with a 6-DoF robot arm.  The mesh of the tree
is obtained by scanning a real dormant tree using an RGB-D
sensor.  To prune a branch, the end effector of the robot
arm must reach the base of that branch in a perpendicular
direction, as demonstrated by the example pruning pose in
Figure \ref{fig:query}.  This benchmark consists of 13
queries in total and each planner is tasked to solve the
query 10 times with a maximum allowable time of 5s.
Interestingly, the thin spindly trees would present
challenges in the form of non-convex cost surfaces for any     
optimization-based planner, including STOMP and cRMPD.  In
the case of cRMPD, our experiments have shown that when the
EGD begins with a naive middle point initialization, namely
$p_m = (p_s + p_g) / 2$, subsequent middle points often
descend quickly into a local minimum that is not
collision-free, resulting in a failure of the planner.
Along similar lines, the consequence of this non-convexity
is also reflected in the low success rate of STOMP, shown in
Table \ref{tb:pruning}.  To address this shortcoming, a
high-quality initialization is of crucial importance.
Therefore, before the invocation of
$CostAwareFreeStateSampler$, cRMPD first takes $K$ samples
around the middle point and compute their costs.
Subsequently, the seed point $p_m$ that is being passed to
the $CostAwareFreeStateSampler$ call is the one with the
lowest cost among the $K$ samples.  

The greedy initialization strategy of cRMPD as described
above has been shown to work effectively.  As one can tell
from the results in Table \ref{tb:pruning}, the trade-off
between time efficiency and path quality is evident. Whereas
STOMP offers the best path quality, it is prone to failure
and entails a large computational burden.  The opposite of
this conclusion holds true for RRT-Connect.  On the other hand,
cRMPD offers us a better trade-off between path quality
and computational burden.  Not only can cRMPD answer the
queries with a 100\% success rate, it also achieves near
optimum performance in terms of planning time and path
length.

\subsection{Limitations}

Despite out-performing the other well-known planners in our
comparative study, cRMPD has some limitations of its own.
First, if a middle point is chosen poorly, it will cause subsequent middle points to diverge from the optimal path.
Since cRMPD is a single-path planner, a middle point, once chosen, can no longer be changed in later iterations and remains in the final path.
We believe that this
shortcoming could be addressed by using cRMPD only as a local
planner in a probabilistically complete top-level planner,
such as RRT.
Furthermore, launching
multiple instances of cRMPD simultaneously in a
multi-threaded fashion also would help get around the
difficulties that may arise with the possible suboptimality of the middle points.

\section{Conclusion}
\label{sec:con}

Best known randomized sampling based algorithms for path
planning, while possessing the highly desirable property of
probabilistic completeness, unfortunately tend to carry out
unnecessary randomized explorations in open spaces and
branch out slowly in narrow passages.  Our ``local'' planner
RMPD has the ability to efficiently bypass local obstacles
using inexpensive resampling, which accelerates excursions
into narrow passages through a divide-and-conquer strategy.
Additionally, cRMPD, our cost-aware version of RMPD, takes
advantage of estimated gradient descent to produce a
cost-optimal middle point.  cRMPD uses a Monte Carlo
sampling strategy where the current sampling distribution is
constantly steered to low-cost samples in the previous
iteration.  Our experimental results demonstrate that cRMPD
possesses superior qualities with regard to time efficiency
and path quality.  This makes cRMPD a powerful new approach
to path planning.

\section{Acknowledgement}

The authors would like to thank the anonymous reviewers for
their insightful feedback.  This project was supported by
the USDA Specialty Crop Research Initiative (SCRI).


\bibliographystyle{aaai}
\bibliography{citations}
\end{document}